\documentclass[runningheads]{llncs}
\usepackage[T1]{fontenc}
\usepackage{xcolor}
\usepackage{graphicx}
\usepackage{amsmath}
\usepackage{amssymb}
\usepackage{booktabs}
\usepackage{pifont}
\newcommand{\cmark}{\ding{51}}%
\newcommand{\xmark}{\ding{55}}%
\usepackage{multirow}
\newcommand{\etal}{\textit{et al.}}
\usepackage{caption}
\usepackage{subcaption}
\usepackage{lscape}
\usepackage{adjustbox}
\usepackage{wrapfig}
\begin{document}
%
\title{2by2: Weakly-Supervised Learning for  \\ Global Action Segmentation}
%
%
\author{Elena Bueno-Benito \orcidID{0009-0006-7566-9771} \and Mariella Dimiccoli \orcidID{0000-0002-2669-400X} }
\authorrunning{E. Bueno-Benito and M. Dimiccoli}
%
\institute{Institut de Robòtica i Informàtica Industrial, CSIC-UPC \\Llorens i Artigas 4-6, 08028 Barcelona, Spain \\ \email{\{ebueno, mdimiccoli\}@iri.upc.edu}\\
}
\maketitle              
\begin{abstract}
This paper presents a simple yet effective approach for the poorly investigated task of global action segmentation, aiming at grouping frames capturing the same action across videos of different activities. Unlike the case of videos depicting all the same activity, the temporal order of actions is not roughly shared among all videos, making the task even more challenging. 
We propose to use activity labels to learn, in a weakly-supervised fashion, action representations suitable for global action segmentation. 
For this purpose, we introduce a triadic learning approach for video pairs, to ensure intra-video action discrimination, as well as inter-video and inter-activity action association.
For the backbone architecture, we use a Siamese network based on sparse transformers that takes as input video pairs and determine whether they belong to the same activity. The proposed approach is validated on two challenging benchmark datasets: Breakfast and YouTube Instructions, outperforming state-of-the-art methods.



\end{abstract}

\begin{figure*}[t]
\centering
 
\includegraphics[trim = 40mm 40mm 40mm 40mm,  clip, width=1.0\linewidth]{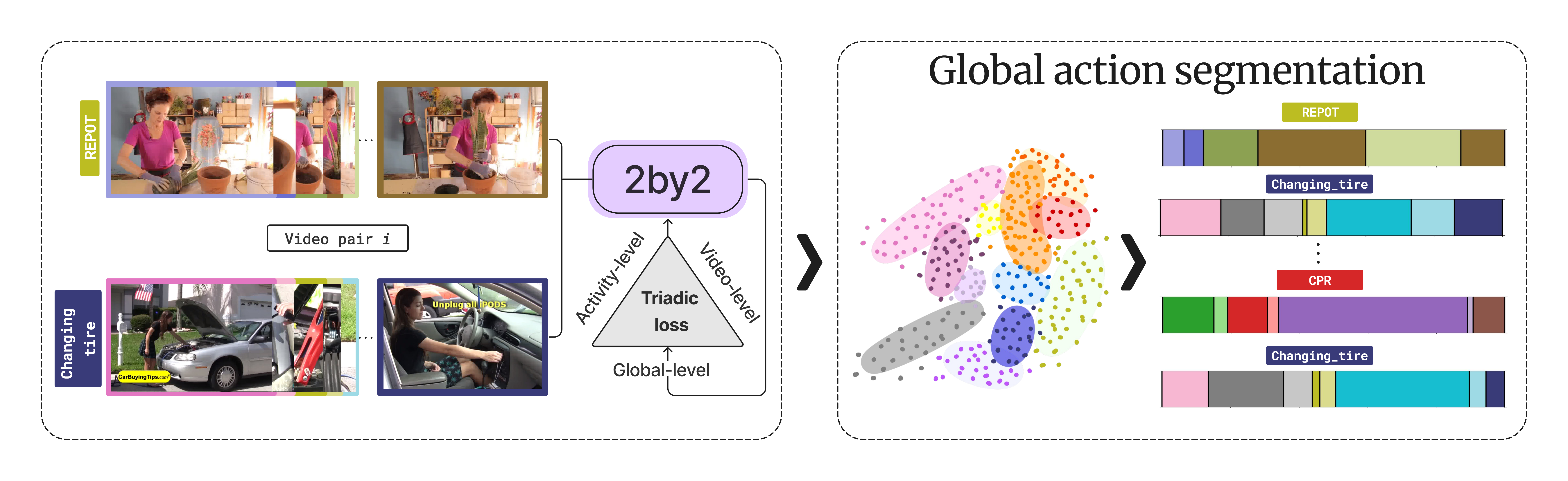}
 \vspace{-2em}
\caption{Our approach compares video pairs through a Siamese network by using binary labels indicating if the videos belong to the same activity or not. We propose a triadic loss function modelling intra-video discrimination, inter-video and inter-activity associations for clustering actions across videos of different activities.}
 \label{fig:overview} 
  \vspace{-2em}
\end{figure*}

\vspace{-2.0em}
\section{Introduction}
\vspace{-0.5em}
Action segmentation, the task of classifying each frame of an untrimmed video plays a fundamental role in various applications such as video surveillance, sports analysis, and content-based video retrieval \cite{he2024,zuckerman2024}. Recently, this task has received significant attention from the research community. The most reliable approaches for action segmentation are fully supervised methods, which require expensive data annotations \cite{Bahrami2023,behrmann2022,Farha2019,Li2020,Lu_2024,Yi2021}. The need for more scalable and practical solutions has led to an increasing interest in developing weakly-supervised \cite{Chang2021,Lu2021,Lu2022,Bin2021,Souri2022,Xu_Weak2024,Zhang2023} and unsupervised techniques \cite{Bueno-Benito2023,Dias2018,Dimiccoli2021,Ding2021,Kukleva2019,Kumar2022,Li2021,Li2024,Sarfraz2021,Sener2018,Tran-Mehmood2023,VidalMata2021,Wang2022,Xu2024}.

Weakly-supervised methods learn to partition videos into action segments using only transcript annotations for each video, typically in the form of actions transcripts (ordered lists of actions) or action sets (unique actions derived from narrations, captions or meta-tags) \cite{Lu2022,Souri2022,Xu_Weak2024,Zhang2023}. This weakly-supervised paradigm contrasts with unsupervised methods, broadly categorized into three types, depending on the matching objective \cite{ding2023survey}: video-level, activity-level, and global-level. 
Video-level segmentation methods aim to segment a single video sequence into distinct actions without considering the relationships between actions in different videos \cite{Bueno-Benito2023,Zexing2022,Li2024,Sarfraz2021,Xu2024}. While they can be effective for practical applications requiring to segment isolated videos one by one, they fail to generalize actions across different videos. Instead, activity-level segmentation methods focus on matching actions across videos that depict the same complex activity \cite{Ding2021,Kukleva2019,Kumar2022,Li2021,Tran-Mehmood2023,Xu_Weak2024}. These methods generally perform poorly at video-level unless temporal smoothing within segments is explicitly modelled. In addition, as they assume or estimate a transcript for each video or set of videos belonging to the same activity,  their generalization ability to other activities is hampered. 
Only Ding \etal \cite{Ding2021} directly addressed global-level segmentation railing on complex activity labels to help discover the constituent actions; however, they do not explicitly model the alignment of actions across videos of the same activity. 

In this paper, we propose a strategy to discover actions across various complex activity videos, offering a broader and more generalized understanding of actions.
Our approach does not require knowledge of video transcripts, but only binary labels indicating whether each pair of videos belongs to the same activity. Therefore, as a weakly-supervised method, it occupies a unique position in the spectrum of action segmentation methods.

\textbf{Our solution}, depicted in Figure \ref{fig:overview}, aims to enhance the clustering of actions in videos on a global scale through the implementation of a Siamese network based on transformers. This network is designed to address the task of determining whether two videos depict the same activity. Instead of using a standard cross-entropy loss, we propose a triadic loss function capturing the temporal dynamics within individual videos, between similar videos, and across various activities. Our contributions are as follows:

\begin{enumerate} 
    \item We propose a novel weakly-supervised framework for the task of global action segmentation that relies on binary activity labels to discover action clusters across videos of different activities. 
    \item We introduce a transformer-based Siamese architecture, that takes input pairs of videos, determines if they belong to the same activity or not and aligns them temporally if predicts that they depict the same activity.
    \item We introduce a triadic loss function that models intra-video action discrimination at the video-level, inter-video and inter-activity action associations at activity and global-level respectively, for robust action understanding. 
    \item We achieve state-of-the-art results on the \textit{Breakfast (BF)} and \textit{Inria Instructional Videos (YTI)} benchmark datasets, demonstrating the method's effectiveness and generalization ability across activities.
\end{enumerate}

\vspace{-1.5em}
\section{Related work}

\label{sec:sota}
\vspace{-0.5em}
\subsection{Action Segmentation}
For a comprehensive and recent survey on temporal action segmentation tasks, readers are referred to \cite{ding2023survey}.
\vspace{-1.0em}
\subsubsection{Supervised Action Segmentation.} Supervised approaches have seen significant advancements over recent years \cite{Bahrami2023,behrmann2022,Farha2019,Li2020,Lu_2024,Yi2021}. Recently, UVAST \cite{behrmann2022} integrates fully and timestamp-supervised learning paradigms via sequence-to-sequence translation. This method refines predictions by aligning frame labels with predicted action sequences. LTContext \cite{Bahrami2023} iterates between windowed local attention and sparse long-term context attention, effectively balancing computational complexity and segmentation accuracy. Lastly, FACT \cite{Lu_2024} performs temporal modelling at both frame-level and action-level, facilitating bidirectional information transfer and iterative feature refinement. However, being fully supervised, all these methods are not scalable and not suited for real applications. 
\vspace{-1.5em}
\subsubsection{Weakly-Supervised  Action Segmentation.} Weakly-supervised techniques have been developed to reduce the need for large annotated datasets. These approaches typically learn to partition a video into several action segments from training videos only using transcripts or other human-generated information to generate pseudo-labels for training \cite{Lu2021,Lu2022,Bin2021,Souri2022,Xu_Weak2024,Zhang2023}. Transcripts have been shown to outperform action set-based methods, while timestamp-based approaches achieve the best results. This suggests that higher levels of supervision generally lead to better performance. In recent years, DP-DTW \cite{Chang2021} has advanced weakly-supervised segmentation by training class-specific discriminative action prototypes. This method represents videos by concatenating prototypes based on transcripts and improves inter-class distinction through discriminative losses. Some methods leverage machine learning models to infer video segments, such as TASL \cite{Lu2021}. Recently, more efficient alignment-free methods have been proposed. MuCon \cite{Souri2022} learns from the mutual consistency between two forms of segmentation: framewise classification and category/length pairs. POC \cite{Lu2022} introduces a loss function to ensure the output order of any two actions aligns with the transcript. Conversely, ATBA \cite{Xu_Weak2024} propose an approach that incorporates alignment by directly localizing action transitions for efficient pseudo-segmentation generation during training, eliminating the need for time-consuming frame-by-frame alignment. None of these methods explicitly addresses the problem of global action segmentation. 
\vspace{-1.0em}
\subsubsection{Unsupervised Action Segmentation.} Unsupervised approaches have been explored by several studies to eliminate the need for annotations \cite{Aakur2019,Bueno-Benito2023,Dias2018,Ding2021,Zexing2022,Kukleva2019,Kumar2022,Li2021,Li2024,Sarfraz2021,Sarfraz2019,Sener2018,Tran-Mehmood2023,VidalMata2021,Wang2022,Xu2024}. As the estimated clusters, lack of semantic labels, the evaluation process requires finding the Hungarian correspondence between the clusters and the actual action classes. The Hungarian matching can be performed for video-level segmentation \cite{Aakur2019,Bueno-Benito2023,Zexing2022,Li2024,Sarfraz2021,Sarfraz2019}, activity-level segmentation \cite{Ding2021,Kukleva2019,Kumar2022,Li2021,Sener2018,Tran-Mehmood2023,VidalMata2021,Wang2022,Xu2024}, or for a global scope across an entire set of videos \cite{Ding2021,Kukleva2019,Li2021}. Depending on the hierarchical level used, methods aim to improve segmentation through these correspondences. Unsupervised techniques in action segmentation typically involve a two-step process: first, learning action representations in a self-supervised manner, followed by employing clustering algorithms to perform action segmentation, assuming a prior knowledge of the number of clusters.
 

In the realm of video-level action segmentation, LSTM+AL \cite{Aakur2019} introduced a novel self-supervised methodology for real-time action boundary detection. Furthermore, it is worth noticing that clustering approaches based on specific similarity metrics have been relatively under-explored in the field of action segmentation. One such method is TW-FINCH \cite{Sarfraz2021}, which captures spatio-temporal similarities among video frames. This employs a temporally weighted hierarchical clustering algorithm, grouping video frames without the need for extensive pre-training, as it directly operates on pre-computed features that augment the conventional FINCH approach with temporal considerations \cite{Sarfraz2019}. In a similar vein, ABD\cite{Zexing2022} identifies action boundaries by detecting abrupt change points along the similarity chain between consecutive features. 

Action representation learning at the individual video level has also gained interest. TSA \cite{Bueno-Benito2023} proposed a method that focuses on this aspect, employing a shallow neural network trained with a triplet loss and a novel triplet selection strategy to learn action representations. These learned representations can be processed using generic clustering algorithms to obtain segmentation outputs. Lastly, the OTAS framework has emerged, offering an unsupervised boundary detection method that combines global visual features, local interacting features, and human-object relational features, contributing to the evolving landscape of action segmentation techniques \cite{Li2024}.


Some approaches at the activity-level leverage the order of scripted activities, emphasizing the minimization of prediction errors, like CTE \cite{Kukleva2019}. Other works combined temporal embedding with visual encoder-decoder pipelines with visual reconstruction loss \cite{VidalMata2021} or with discriminative embedding loss \cite{Swetha2021}. ASAL \cite{Li2021} explored deep learning architectures, such as ensembles of autoencoders and classification networks that exploit the relationship between actions and activities. CAD \cite{Ding2021} introduced a framework that discovers global action prototypes based on high-level activity labels. One notable aspect of these methods is the recognition that actions in task-oriented videos tend to occur in similar temporal contexts. As a result, strong temporal regularization techniques have been developed to partially obscure visual similarities \cite{Kukleva2019,Sener2018}.  
Recently, optimal transport has gained popularity in unsupervised learning to generate effective pseudo-labels and train for frame-level action classification. TOT \cite{Kumar2022} proposed a joint self-supervised representation learning and online clustering approach that directly optimizes unsupervised activity segmentation using video frame clustering as a pretext task. UFSA \cite{Tran-Mehmood2023} extends TOT by combining frame and segment-level cues to improve permutation-aware activity segmentation. Furthermore, TOT and UFSA use a Hidden Markov Model (HMM) approach to decode segmentations given a fixed or estimated action order, respectively. In contrast, ASOT \cite{Xu2024} proposed a method via optimal transport that yields temporally consistent segmentations without prior knowledge of the action ordering, required by previous approaches. Suitable for both pseudo-labeling and decoding.

Although global-level understanding provides the most comprehensive insight into the relationships between activities and actions in videos, only a few methods have explored training at this level. CAD \cite{Ding2021} is the first work to operate at the highest level of global matching. In CTE \cite{Kukleva2019}, the methods extended their configuration considering all complex activities. Firstly, the protocol executes a bag-of-words clustering on the videos to divide them into multiple pseudo-activities. Subsequently, they perform action clustering within each pseudo-activity individually. In other words, they apply their action segmentation at the activity level within classes of pseudo-activity. Their approach still does not accommodate potential actions shared between activities. ASAL \cite{Li2021} and CAD \cite{Ding2021} present their results aligned with this protocol. 

\vspace{-1.0em}
\subsection{Video Alignment}
Video alignment is a process aimed at synchronizing and matching video sequences for various applications, such as action recognition model creation, behavioural analysis, and multimedia content generation. This field encompasses a range of techniques. Traditionally, methods like Dynamic Time Warping (DTW), Canonical Correlation Analysis, ranking or match-classification objectives, and the differentiable version of DTW, Soft-DTW, have been used to tackle the challenging task of aligning video frames \cite{Galen2023,arandjelovic2017,Cuturi2017,sermanet2018} in videos depicting a same action. Recently, LAV \cite{Haresh2021} have utilized Soft-DTW combined with temporal intra-video contrastive loss to align video frames effectively. Drop-DTW \cite{Dvornik2021}, an extension of DTW, introduces a "trash bucket" to the cost matrix, allowing for the classification of background frames and robust alignment in the presence of outliers. VAVA \cite{Liu2022} employs optimal transport with a bi-modal Gaussian prior and a virtual frame for unmatched frames.

TCC \cite{Dwibedi2019} was the first to introduce cycle-consistency for aligning video frames by maximizing cycle-consistent embeddings between sequences. GTCC \cite{Donahue2024} extends the TCC approach to manage more complex alignment scenarios. However, most of these techniques were developed for general video alignment or related tasks, and their direct application to unsupervised action segmentation has been never explored so far. In this paper, we propose for the first time to leverage video alignment for action segmentation.


\begin{figure*}[t]
    \centering
    \includegraphics[trim = 40mm 40mm 40mm 40mm,  clip, width=1\linewidth]{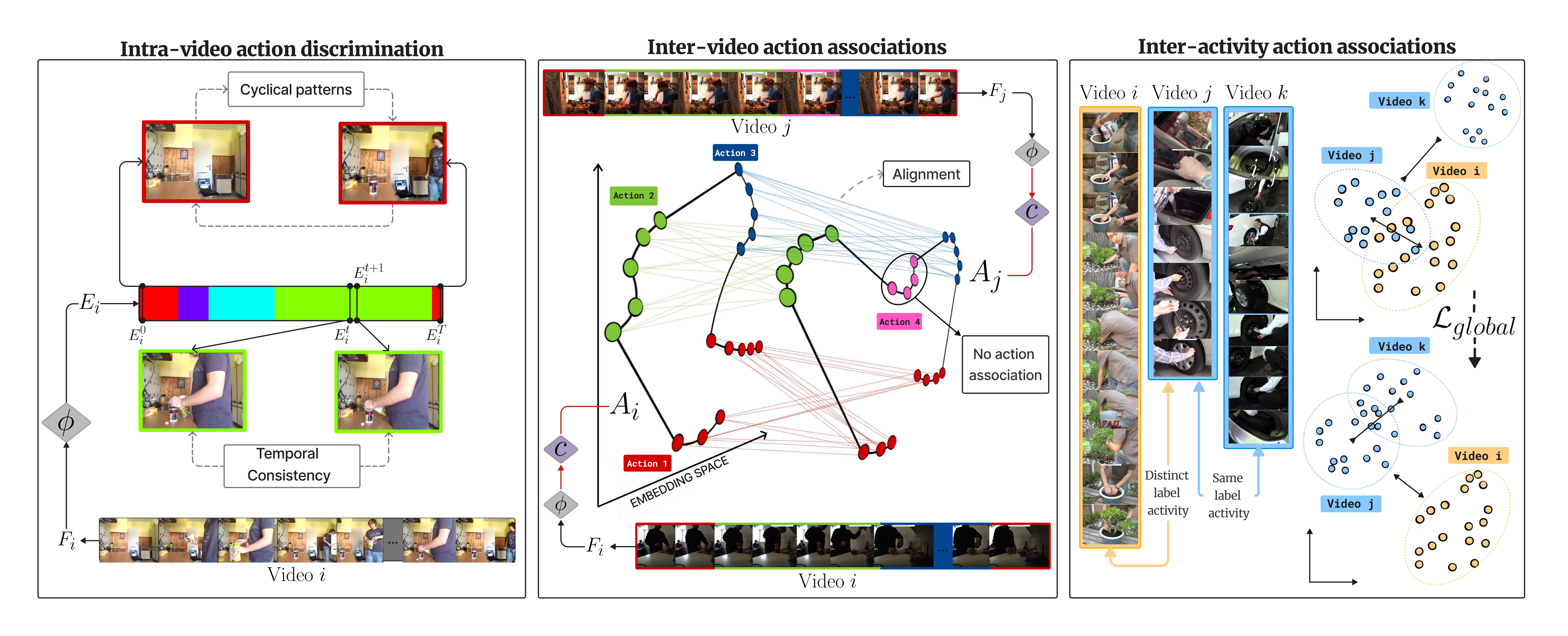}
     \vspace{-1.5em}
    \caption{Overview of the proposed 2by2 framework. The figure illustrates our triadic learning approach: intra-video action discrimination, which enhances cross-temporal consistency within a single video (first box); inter-video action associations, which align action frames among similar videos (second box); and inter-activity action associations, which establish global correspondence between different videos (third box). The red arrows indicate steps specific to the training phase.}
    \label{fig:triadiclearning} 
    \vspace{-1.0em}
\end{figure*}

\section{2by2: Learning Unknown Actions in a Global Manner}

\label{sec:method}
This section presents a weakly-supervised, triadic action learning approach for global action segmentation (see Figure \ref{fig:triadiclearning}), 
aiming at modeling:
\begin{itemize}
    \item[(i)] \textit{Intra-video action discrimination} (video level): Video frames sharing the same action with their nearest neighbours exhibit temporal consistency. Moreover, actions typically do not occur at the beginning or end of videos. Thus, a video can be interpreted as a cyclic temporal sequence.
     \item[(ii)] \textit{Inter-video action associations} (activity level): For videos categorized under the same activity, segments within these videos exhibit similarity, facilitating the alignment of actions across them.
     \item[(iii)] \textit{Inter-activity action associations} (global level): Videos representing different activities that share common actions should be closer in the representational space compared to those that do not share actions.
\end{itemize}
 \vspace{-1.0em}
\subsection{Problem Formulation}
Given a large set $V$ of complex activity videos from a dataset belonging to $C$ complex activities, each video $v_{i}$ in $V$ is annotated with a complex activity label $a \in [1, C]$. Our objective is to associate each video frame $x_t$, with an action label $n$ from $N$ possible actions. These $N$ actions are constituent steps shared among the $C$ complex activities. For each video $v_i$, we define the feature matrix $F_{i}$, where each row $F_{i}^{t}$ corresponds to an $d$-dimensional feature vector at time $t$ in a video $v_{i}$. 
Given the initial features of a video $F_{i}$, our objective is to learn a parametric function $\phi$ that categorizes video frames into the $N$ possible actions, resulting in embeddings $E_{i}$, obtained as $E_{i} = \phi(F_{i}), \forall v_{i} \in V$.
\subsection{Architecture}
To learn $\phi$, we propose a Siamese architecture that takes as input pairs $(F_{i}, F_{j})$ for all $v_{i}, v_{j} \in V$ with $i \neq j$. 
This architecture consists of two identical LTContext networks \cite{Bahrami2023}, 
specifically designed to capture long-term temporal dependencies,
that work in tandem and compare the similarity between their outputs, denoted as $(E_{i},\;E_{j})$ at the end.   

During training, to ensure that videos sharing the same activity have well-aligned representations, we introduce a context-drop function $c$, inspired by \cite{Donahue2024}. This function is designed to handle background and redundant frames by enforcing multi-cycle consistency for alignable embeddings and poor alignment for droppable embeddings. The context-adjusted embeddings are calculated as $A_{i} = c(E_{i}), \forall v_{i} \in V$.

\subsection{Triadic Loss}

\subsubsection{Intra-video Discrimination Loss. }
The output of $\phi$ at different stages, denoted as $\phi_s$, is used to calculate the loss at video level, enhancing the model's ability to learn fine-grained temporal structures. We incorporate a mean squared error smoothing loss, as introduced by \cite{Farha2019} and used in \cite{Bahrami2023,Li2020,Yi2021}. Considering that actions occurring in an activity video should be temporally contiguous, this loss is applied to the per-frame actions to alleviate over-segmentation. Moreover, we also propose a cyclic variant, based on the assumption $(i)$ described at the beginning of Section \ref{sec:method}. Specifically, this variant compares the embeddings at the end of the output sequence with those at the start, across different stages of the feature extraction network $\phi$. This is driven by the fact that actions often exhibit cyclical patterns in videos. Mathematically, our video-level loss is defined as follows:
\begin{multline} 
\label{video}
\mathcal{L}_{video} (i)  = \frac{1}{|S||T+1|}   \Big ( \sum_{s} \sum_{t}\left |  \log  \phi_{s}(F_{i}^{t+1}) - \log  \phi_{s}(F_{i}^{t+1}) \right | \\
+ \left |  \log  \phi_{s}(F_{i}^{T})   - \log  \phi_{s}(F_{i}^{0}) \right |   \Big ),
\end{multline} 
where $T$ is the total number of frames and $S$ is the number of stages in $\phi$ in a video $i, \,\forall v_{i} \in V$.

\subsubsection{Inter-video Associations Loss.}
For segment-level learning, we adopt the GTCC loss function proposed by \cite{Donahue2024}, denoted as $\mathcal{L}_{activity}$, to synchronize frames of videos depicting the same activity. We utilize context-adjusted embeddings $A{i}$ generated by our context-drop function layer $c$. Specifically, for each pair $v_{i}, v_{j} \in V$ of videos, GTCC computes the probability of dropping $v^{t}_{i}$ given $v_{j}$ for all $t \in T$ using the function $c$. The loss function is defined as:
\begin{equation}
    \mathcal{L}_{GTCC}(v_{i}|v_{j}) = \sum_{t} \Big ((1- \text{P}_{\text{drop}}(v^{t}_{i}|A_{j}))\cdot \mathcal{L}_{multi-cbr} + \frac{\text{P}_{\text{drop}}(v^{t}_{i}|A_{j})}{ \mathcal{L}_{multi-cbr}}\Big ),
\end{equation}
where $ \mathcal{L}_{multi-cbr}$ is a multi-cycle back regression loss, and $\text{P}_{\text{drop}}(v^{t}_{i}|A_{j})$ is the probability of dropping each video frame $v^{t}_{i}$ given $A_{j}$ (refer to \cite{Donahue2024} for more details). Our activity loss, $ \mathcal{L}_{activity}$ is defined as the sum of $GTCC$ loss of $v_{i}$ given $A_{j}$ and vice-versa. This loss leverages the principle of Temporal Cycle Consistency (TCC) \cite{Dwibedi2019}, ensuring that corresponding frames in videos with identical action sequences are closely aligned in the feature space. This approach addresses variations in action order, redundant actions, and background frames, thereby enhancing the quality of video representations. To the best of our knowledge, this marks the first application of video alignment for temporal action segmentation.
 \vspace{-1.0em}
\subsubsection{Inter-activity Associations Loss.}
We learn the global representation of a video clip by using a contrastive loss. We employ contrastive learning to minimize the distance between videos of the same activity while maximizing the distance between videos of different activities. This ensures that videos depicting the same activity are closer in the feature space than videos that are not. The global contrastive loss has the following formulation:    
\begin{equation}
\label{global}
\mathcal{L}_{global}(i, j) = (1 - y) \cdot d(E_{i}, E_{j}) + y \cdot \max(0, m - d(E_{i}, E_{j}))
\end{equation}
where $d(E_{i}, E_{j})$ denotes the distance between the representations $E_{i}$ and $E_{j}$ obtained by $\phi$, and $y \in \{0, 1\}$ is a binary value such that $y = 0$, if the two videos belong to the same activity ($a_i = a_j$), and $y = 1$, if they belong to different activities ($a_i \neq a_j$). The margin $m$ ensures sufficient separation between videos of different activities. The term $(1 - y) \cdot d(E_{i}, E_{j})$ minimizes the distance for videos of the same activity, while $y \cdot \max(0, m - d(E_{i}, E_{j}))$ maximizes the distance for videos of different activities by pushing them apart by at least the margin $m$.

\subsubsection{}

The combined loss function that governs the training for all pair videos $\{v_{i},v_{j}\} \in V$ of our model is formulated as:

\begin{equation}
\label{total_loss}
\begin{aligned}
\mathcal{L}_{\text{train}}(\phi, c)) = \begin{cases} 
\alpha \mathcal{L}_{\text{global}}(i, j) + (1 - \alpha) \mathcal{L}_{\text{activity}}(i, j) \\
\,\,\,\,\,\,\,\,\,\,\,\,\,\,\,\,\,\,\,\,\,\,\,\,\,\,\quad + \beta (\mathcal{L}_{\text{video}}(i) + \mathcal{L}_{\text{video}}(j)), & \text{if } v_i = v_j \\ 
\mathcal{L}_{\text{global}}(i, j) + \beta (\mathcal{L}_{\text{video}}(i) + \mathcal{L}_{\text{video}}(j)), & \text{if } v_i \neq v_j 
\end{cases}
\end{aligned}
\end{equation}
where $\alpha$, and $\beta$ are hyperparameters that balance the contributions of the global, activity, and video loss components. Incorporating this loss in our model allows us to leverage the weak supervision effectively, making the clustering of video frames more discriminative and improving the overall performance of action segmentation and classification tasks in a global manner.

 \vspace{-1.0em}
\section{Experimental Setup}
 \vspace{-0.5em}
\subsubsection{Datasets.} We present results on two well-known datasets used for temporal action segmentation:
\textbf{Breakfast Action Dataset (BF)} \cite{Kuehne2014} is one of the largest fully annotated collections available for temporal action segmentation. It includes 1712 videos, featuring 10 activities related to breakfast preparation. These activities are performed by 52 individuals across 18 different kitchens. Each video has an average of 2099 frames. Remarkably, only 7\% of the frames are background frames. \textbf{Youtube INRIA Instructional Dataset (YTI)} \cite{Alayrac2016} includes 150 instructional videos from YouTube, covering 5 different activities such as changing a car tire, preparing coffee, and performing cardiopulmonary resuscitation (CPR). The videos have an average duration of 2 minutes. A significant challenge with this dataset is the high proportion of background frames, which make up 63.5\% of the total frames.
  \vspace{-0.5em}
  
\subsubsection{Features.}
To ensure a fair comparison with related work, we utilized the same input features as recent methods. For the BF dataset, we used the IDT features \cite{Wang2013} provided by the authors of CTE \cite{Kuehne2014} and SCT \cite{Rohrbach2016}. These features capture motion information by tracking dense points in the video and computing descriptors such as Histogram of Oriented Gradients, Histogram of Optical Flow (HOF)\cite{laptev2008}, and Motion Boundary Histogram. Additionally, for further comparison in the BF dataset, we employ I3D features \cite{carreira2017} extracted from the Inflated 3D ConvNet, which leverages both spatial and temporal convolutions to learn video representation. For the YTI dataset, we use the same features as \cite{Alayrac2016,Ding2021}. These 3000-dimensional feature vectors are formed by concatenating HOF descriptors with features extracted from the VGG16-conv5 layer \cite{simonyan2014}. 
 \vspace{-0.5em}
 \subsubsection{Metrics.} To evaluate the performance of our temporal action segmentation methods, we employ 1) Mean over Frames (MoF), which calculates the accuracy as the mean percentage of correctly classified frames across all videos, providing a direct indication of overall segmentation performance; 2) F1-Score, which is the harmonic mean of precision and recall, accounting for both false positives and false negatives. Precision is the ratio of correctly predicted action frames to the total predicted action frames, while recall is the ratio of correctly predicted action frames to the total actual action frames; 3) MoF with Background (MoF-BG), which calculates the accuracy considering both action and background frames, essential for understanding how well the segmentation method distinguishes between action and non-action frames, especially given the high proportion of background frames in the YTI dataset. To enable direct comparison, we follow the procedure used in previous work \cite{Bueno-Benito2023,Li2024,Kukleva2019,Zexing2022,Sener2018}, reporting results by removing the ratio ($\tau = 75\%$) of the background frames from the video sequence.
  \vspace{-0.5em}
\subsubsection{Evaluation Setting.}
In our study, we adopt the global evaluation methodology proposed by \cite{Kukleva2019}. This methodology involves grouping videos into coherent subsets $K$ and representing them using a bag-of-words (BoW) approach. These representations are then clustered into $K'$ groups of pseudo-activities and $K$ subgroups of actions are inferred. Each video is temporally segmented by assigning each frame to one of the ordered groups using the Viterbi decoder. A background model is introduced to deal with irrelevant segments. Throughout the results of this work, the inclusion of BoW and Decoding refers to the integration of the aforementioned global inference process, which we will refer to as the \textit{post-processing protocol}.

For evaluation, we perform a Hungarian matching between the inferred clusters and the ground-truth labels to compute the metrics. Specifically, we assume in the case of the Breakfast dataset $K'=10$ activity clusters with $K=5$ sub-actions per cluster. Subsequently, we match 50 different sub-action clusters with 48 ground-truth sub-action classes, with frames of the leftover clusters set as background. Finally, we assess the accuracy of the unsupervised learning configuration on the YouTube Instructions dataset, employing $K'=5$ and $K=9$, subsequently matching $45$ distinct sub-action clusters with $47$ ground-truth sub-action classes.
 \vspace{-0.5em}

\subsubsection{Training Details.}
 To ensure that each video in our training set has at least one pair from the same activity and one pair from a different activity, we construct the training set by including all possible combinations of videos belonging to the same activity. Since segment-level learning requires a strong initialization to align actions between videos, we adopt a two-stage training approach. Initially, the model is trained with global-level and video-level modules using Eq. \ref{video} and \ref{global}, respectively. Subsequently, the model is used to initialize the second stage, where it is trained using the full loss function in Eq. \ref{total_loss}. In a stratified fashion, we select a subset of pairs from different activities, ensuring an equal number of same-activity and different-activity pairs. Given a large number of combinations, in each epoch, we take a batch including 50\% of the dataset of possible pairs for each epoch. Note that each epoch uses a batch size of 32 pairs for the BF dataset and 8 pairs for the YTI dataset. We simultaneously train a 4-layer feed-forward neural network for the drop-context function, $c$, along with $\phi$. To enhance computational efficiency, we down-sample all videos to 256 frames per video by randomly removing frames distributed throughout each video, similar to \cite{Kumar2022,Tran-Mehmood2023}. This technique reduces frame redundancy and ensures that the frames represent the entire video. We use the same parameters as specified in \cite{Bahrami2023} and \cite{Donahue2024} for each network. The training process employs the ADAM optimizer, with a learning rate of $2e^{-4}$ and a weight decay of $10^{-4}$. For the parameters $\alpha$ and $\beta$, we select the values $0.15$ and $0.5$, respectively.  
\vspace{-0.5em}

\subsection{Comparative methods.}

The method more similar to ours in terms of scope, i.e. global action segmentation, and information used, i.e. activity labels, is CAD \cite{Ding2021}. For the sack of completeness, we compute results with a global matching scope of state-of-the-art methods conceived for action segmentation at activity level. These include on the one side unsupervised methods such as ASOT \cite{Xu2024}, CTE \cite{Kukleva2019} and ASAL \cite{Li2021} that train a network for each activity hence using our same pseudo-labels; on the other side, they include weakly-supervised methods such as ATBA \cite{Xu_Weak2024} that instead use a transcript for each video, resulting in a much stronger level of supervision.

 \vspace{-1.5em}

\section{Results}

\subsection{Comparison with the State-of-the-art}

 \vspace{-0.5em}

\begin{table}[t]
    
    \centering
    \begin{minipage}{0.48\textwidth} 
     \resizebox{1.05 \textwidth}{!}{ 
     
    \centering
    \begin{tabular}{l | l | c | c| c| c  c  } 
        \toprule
          \multicolumn{7}{c}{\textbf{BF}}  \\ [0.5ex] 
      \hline
      \hline  
    \textbf{Supervision}  & \textbf{Approach} & \textbf{F} & \textbf{BoW} &  \textbf{D}  &   \hspace*{0.1cm}\textbf{F1}  &  \textbf{MoF}   \\  
    \hline  
    \multirow{5}{*}{Unsupervised} & CTE \cite{Kukleva2019} &\hspace*{0.1cm}IDT\hspace*{0.1cm}& \cmark & \cmark & - & 18.5  \\
    & ASAL \cite{Li2021}&IDT& \cmark & \cmark &  - & 20.2   \\ 
    \cline{2-7}
    & ASOT* \cite{Xu2024} & IDT & \cmark & \cmark  & 20.2 & 21.6  \\ 
    \hline 
     \multirow{5}{*}{Weak} & \multirow{3}{*}{CAD \cite{Ding2021}}  & \multirow{3}{*}{IDT}& \xmark & \xmark  &  - & 10.9 \\
    &  & & \cmark & \xmark  &  - & 17.7 \\
    &   & & \cmark & \cmark &  - & \underline{23.4}  \\  
    \cline{2-7}
    & \color{blue} 2by2  &IDT   & \cmark & \cmark &  \textbf{20.6} & \textbf{24.6}   \\ 
    
    \hline 
    \multicolumn{7}{c}{} \\
    \toprule  
    Unsupervised \hspace*{0.5cm} &  ASOT* \cite{Xu2024}  & I3D  & \cmark & \cmark & 16.9  & 18.1     \\
    \hline 
    Weak-transcripts \hspace*{0.5cm} & ATBA* \cite{Xu_Weak2024} & I3D & \cmark & \cmark & 20.0 & 17.7 \\
    \hline 
    \multirow{2}{*}{Weak-activity labels \hspace*{0.5cm}} & CAD \cite{Ding2021} &I3D& \xmark & \xmark &  - & 19.2 \\  
    &  \color{blue} 2by2\hspace*{0.1cm}& I3D & \cmark & \cmark &\textbf{17.5} & \textbf{20.7}   \\
    \hline 
    \end{tabular}     }

    \end{minipage}
    \hfill
    \begin{minipage}{0.48\textwidth} 
    \resizebox{1.05\textwidth}{!}{ 
    \centering
    
    \begin{tabular}{l | l | c | c | c c c   } 
    
        \toprule 
     \multicolumn{7}{c}{\textbf{YTI}}  \\ [0.5ex] 
        \hline \hline 
    \textbf{Supervision} & \textbf{Approach}  & \textbf{BoW} & \hspace*{0.1cm}\textbf{D}\hspace*{0.1cm}  & \hspace*{0.1cm} \textbf{F1} & \textbf{MoF}&  \textbf{MoF\tiny{-BG}} \\  
    \hline  
     \multirow{2}{*}{Unsupervised } &CTE \cite{Kukleva2019}  & \cmark & \cmark  & -&19.4  & 10.1 \\
     \cline{2-7}
     & ASOT* \cite{Xu2024} & \cmark & \cmark  &  15.26 & 18.6 & 9.9 \\

    \hline 
    
    \multirow{2}{*}{Weak }& CAD \cite{Ding2021} & \xmark & \xmark   & 12.10 & 15.7 & -\\  
    &  \color{blue} 2by2   & \cmark & \cmark    & \textbf{16.53}& \textbf{23.6} & \textbf{11.4} \\
    
     \hline
     
    \end{tabular}    } 
    
    \vspace{6.2em}
   
     \end{minipage}
      \vspace{1em}
      \caption{Action Segmentation results on the BF and YTI datasets by applying the Hungarian matching at global-level.  The dash indicates "not reported."  (*) denotes results computed by ourselves. "F" denotes the type of features used. "D" indicates the use of Viterbi decoding. Both marks denote evaluation as per \cite{Kukleva2019}. The best results are marked in bold.}
     \vspace{-2em}
     \label{tab:comparision_sota} 
\end{table}

\subsubsection{Breakfast dataset (BF).} The results obtained by using the IDT features as input demonstrate a consistent performance improvement over prior methods (refer to left-hand table \ref{tab:comparision_sota}). We achieved a $+1.2\%$ improvement in MoF with respect to CAD, highlighting the efficacy of our global training approach with binary labels. 

We computed the results at the global level of ASOT \cite{Xu2024}, by following the evaluation protocol described above. 2by2 outperforms it in terms of MoF by $+3\%$ and in terms of F1-score by $+0.4\%$. Similar trends are observed when using I3D features as input. Compared to state-of-the-art methods, the 2by2 framework proves effective regarding MoF and F1-score. ATBA \cite{Xu_Weak2024} exhibits a higher F1-score but a lower MoF than 2by2, likely due to its use of transcripts for each video, providing stronger supervision with respect to our method but poorer generalization across activities. This could be attributed to the fact that these methods were not specifically designed for global training, highlighting the critical importance of inter-activity learning which is currently lacking in other unsupervised methods.

\begin{table}[t]

    \centering
    \begin{minipage}{0.45\textwidth}
        \centering
        \begin{tabular}{c c c | c    } 
              \multicolumn{4}{c}{\textbf{YTI}} \\ [0.5ex] 
            \hline
            \hspace*{0.1cm} $ \mathcal{L}_{\text{video}}$\hspace*{0.1cm}  & \hspace*{0.1cm} $\mathcal{L}_{\text{activity}}$\hspace*{0.1cm}  &  $\mathcal{L}_{\text{global}}$  &  \textbf{MoF} \hspace*{0.25cm} \\ 
            \hline 
            \cmark & \cmark & \cmark & \textbf{23.6}    \\
            \xmark & \cmark & \cmark & 21.9   \\
            \cmark & \xmark & \cmark & 22.5  \\
            \xmark & \xmark & \cmark & 21.8   \\
            \hline 
            \multicolumn{4}{c}{} \\
            \hline  
            \multicolumn{3}{l|}{\textit{Base}} &  \textbf{23.6}\\
            \multicolumn{3}{l|}{No $k\_$means init} & 21.1   \\
            \multicolumn{3}{l|}{No cycled MSE} &  21.0    \\
            \multicolumn{3}{l|}{ \footnotesize{No $k\_$means init and cycled}} &  20.4   \\
            \hline  
            \end{tabular}
        \vspace*{0.1cm}
        \caption{Ablation studies on the YTI dataset, highlighting the importance of the three loss terms, as well as of the concept of temporal cycles and the initialization with k-means.}
        \label{tab:ablation_study} 
    \end{minipage}
    \hfill
    \begin{minipage}{0.52\textwidth}
        \centering 
        \vspace{0.1cm}
        
        \hspace{-0.75cm}
        
        \includegraphics[trim = 70mm 40mm 48mm 40mm,  clip, width=1.0\textwidth]{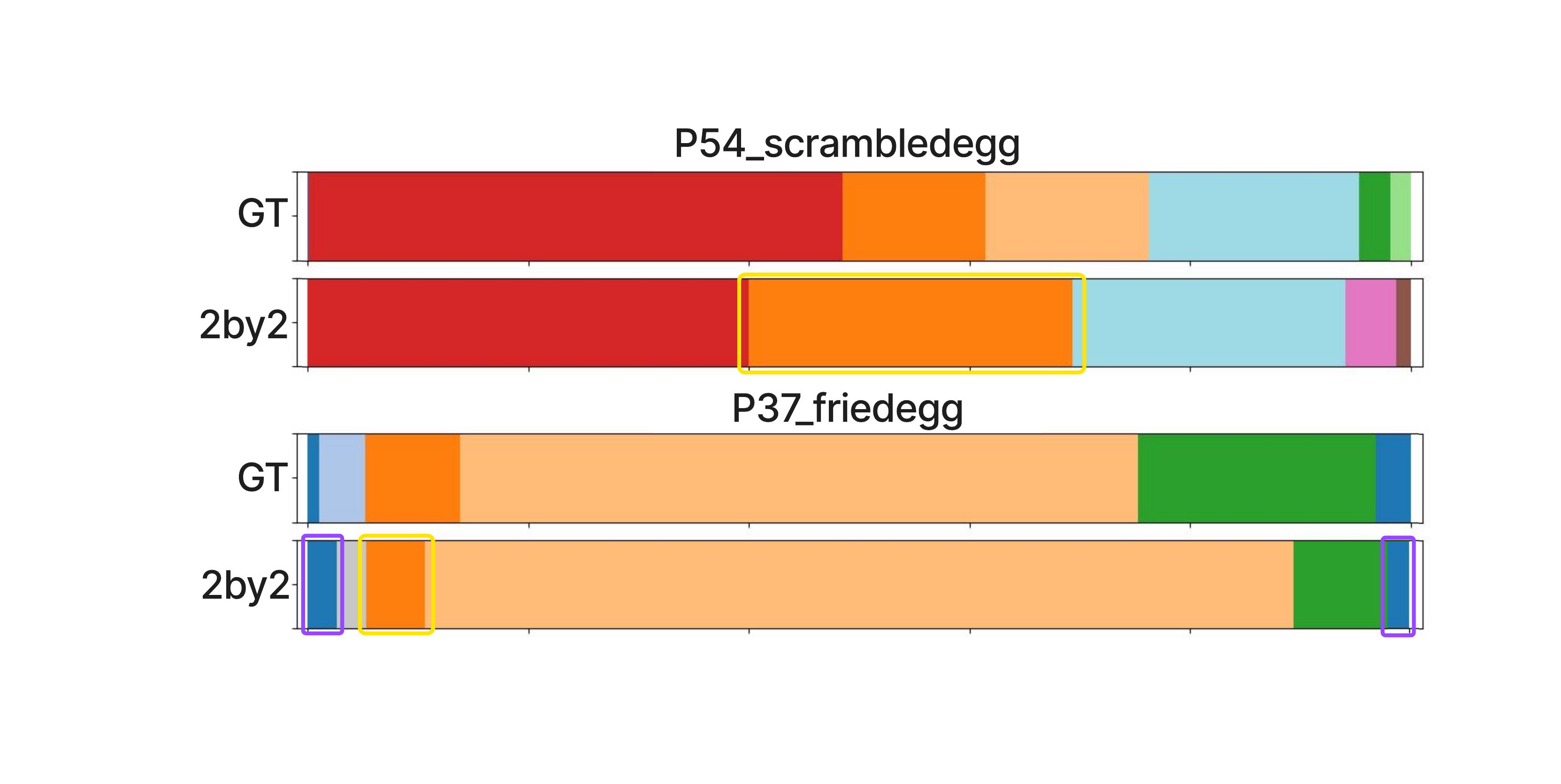}
        \vspace{0.1cm}
        \captionof{figure}{Examples from BF ("scrambled egg" and "fried egg" activities). Comparison of ground truth (GT) segmentation and our 2by2 framework. 2by2 discovers common action steps across activities (see yellow segments) and captures the cyclic nature of the videos (see purple segments). }
        \label{fig:example}
      
    \end{minipage}
    \vspace{-2.5em}
\end{table}

\subsubsection{Inria Instructional Videos (YTI).} The performance of our 2by2 framework also shows marked improvements over previous methods on the YTI (refer to right-hand table \ref{tab:comparision_sota}). We achieve an increase in MoF of $+4.2\%$ without background and $+1.3\%$ with background. This improvement in the F1 score is likely attributed to the non-repetitive nature of actions within activities in this dataset. Our 2by2 framework effectively enhances segmentation accuracy compared to ASOT, the leading unsupervised activity-level segmentation method. Similar to BF, our results underscore the effectiveness of inter-activity training. Furthermore, leveraging global-level training with CAD, we observe significant improvements of $+7.9\%$ in MoF and $+4.4\%$ in F1 score.

The observed performance improvements in both datasets are likely due to the framework's ability to identify better shared actions among pseudo-activity classes caused by inaccurate pseudo-labels and the enhanced initialization of the Bag of Words (BoW) model through video alignment.
 

 
\vspace{-0.5em}
 
\subsubsection{Qualitative Result.} 
\vspace{-0.5em}
In Fig.~\ref{fig:example}, we observe examples closely aligning with the ground truth segments, accurately capturing both large and small segments. The enhanced segmentation arises from multi-level processing within our framework. The activity-level component (GTCC) facilitates precise segment alignment, while the global aspect improves activity differentiation and reduces misclassification. At the video level, our framework maintains temporal consistency and cyclic patterns, reducing over-segmentation and enhancing alignment.
\vspace{-0.5em}
\subsection{Ablation study} 
In Table \ref{tab:ablation_study}, we show the importance of modelling all three levels of learning, by using $ \mathcal{L}_{\text{video}}$, $\mathcal{L}_{\text{activity}}$ and $\mathcal{L}_{\text{global}}$. Specifically, we observe that the elimination of the intra-video component significantly impacts our method's performance, highlighting the detrimental effect of relying solely on the global loss. Additionally, since the inter-video component is introduced in the second stage, it becomes clear that robust initialization in the first stage is essential for $\mathcal{L}_{\text{activity}}$ to effectively guide the alignment and segmentation processes. This underscores that the global loss alone in the first stage is insufficient for achieving optimal performance.

Furthermore, we ablate the effect of initializing the activity cluster for the last layer used for $\mathcal{L}_{\text{global}}$ by using k-means instead of random initialization. Additionally, the negative impact of removing the cyclic component from $\mathcal{L}_{\text{video}}$ is evident.


\vspace{-1em}
 
\section{Conclusion}
\vspace{-0.5em}
This paper introduced 2by2, a novel framework for weakly supervised temporal action segmentation in untrimmed videos encompassing different activities. The proposed architecture consists of a Siamese transformer-based network that takes input pairs of videos and determines if they belong to the same activity or not. If they do, the videos are also temporally aligned. A key innovation of our approach is the direct action alignment between videos, crucial for accurately matching corresponding segments. This is enabled by the Siamese two-stage architecture that ensures robust initialization for temporal alignment.
By explicitly modelling intra-video action discrimination, inter-video action associations, and inter-activity action associations, our method significantly outperforms state-of-the-art approaches on the challenging BF and YTI datasets. 

\subsubsection{Acknowledgements} This work was supported by the grant PRE2020-094714, the project \small{PID2019-110977GA-I00} and the project \small{PID2023-151351NB-I00} funded by \small{MCIN/ AEI /10.13039/501100011033}, by "ESF Investing in your future" and by ERDF, UE.
\vspace{-1em}
%
%
%

\bibliographystyle{splncs04}
\bibliography{egbib}

\end{document}